%% file: DTP.tex
\documentclass[letterpaper, 10pt, conference]{template/ieeeconf}  
\input{template/math_commands.tex}

\usepackage[T1]{fontenc}
\usepackage{wrapfig}
\usepackage{booktabs}
\usepackage{amsmath}
\usepackage{amsmath} 
\usepackage{amssymb}  
\usepackage{accents}
\usepackage{todonotes}
\usepackage{colortbl}
\usepackage{caption}
\usepackage{diagbox}

\presetkeys%
{todonotes}%
{inline,backgroundcolor=yellow}{}
\usepackage{chngcntr}
\usepackage{siunitx}
\usepackage{tikz}
\usepackage{adjustbox}
\usepackage{url}
\usepackage{multirow}
\usepackage{subfigure}
\usepackage[ruled,linesnumbered]{algorithm2e}

\presetkeys%
    {todonotes}%
    {inline,backgroundcolor=yellow}{}
\usepackage{chngcntr}
\usepackage{siunitx}
\usepackage{tikz}
\usepackage{adjustbox}
\usepackage{url}
\usepackage{subfigure}
\usepackage{todonotes}
\usepackage{colortbl}
\usepackage{caption}
\usepackage{diagbox}
\usepackage{makecell}
\IEEEoverridecommandlockouts                              

\title{\LARGE \bf
Diffusion Trajectory-guided Policy for Long-horizon \\ Robot Manipulation
}

\newif\ifblind

\ifblind
  \author{Anonymous Authors}
\else
\author{Shichao Fan$^{1}$, Quantao Yang$^{4}$, Yajie Liu$^{2}$, Kun Wu$^{3}$, Zhengping Che$^{3}$, Qingjie Liu$^{2*}$, Min Wan$^{1}$%
\thanks{$^{1}$ Shichao Fan and Min Wan are with School of Mechanical Engineering and Automation, BeiHang University, China. shichaofan@buaa.edu.cn.}%
\thanks{$^{2}$ Yajie Liu and Qingjie Liu are with School of Computer Science and Engineering, BeiHang University, China. *Corresponding Author: qingjie.liu@buaa.edu.cn.}%
\thanks{$^{3}$ Kun Wu and Zhengping Che are with Beijing Innovation Center of Humanoid Robotics, China.\{gongda.wu, z.che\}@x-humanoid.com}%
\thanks{$^{4}$ Quantao Yang is with Division of Robotics, Perception and Learning (RPL), KTH Royal Institute of Technology, Sweden.}%
  }
\fi

\usepackage[colorlinks=true, linkcolor=blue, citecolor=blue, urlcolor=blue]{hyperref}  

\begin{document}

\maketitle
\thispagestyle{empty}
\pagestyle{empty}

\begin{abstract}
Recently, Vision-Language-Action models (VLA) have advanced robot imitation learning, but high data collection costs and limited demonstrations hinder generalization and current imitation learning methods struggle in out-of-distribution scenarios, especially for long-horizon tasks. A key challenge is how to mitigate compounding errors in imitation learning, which lead to cascading failures over extended trajectories. To address these challenges, we propose the Diffusion Trajectory-guided Policy (DTP) framework, which generates 2D trajectories through a diffusion model to guide policy learning for long-horizon tasks. By leveraging task-relevant trajectories, DTP provides trajectory-level guidance to reduce error accumulation. Our two-stage approach first trains a generative vision-language model to create diffusion-based trajectories, then refines the imitation policy using them.
Experiments on the CALVIN benchmark show that DTP outperforms state-of-the-art baselines by $25\%$ in success rate, starting from scratch without external pretraining. Moreover, DTP significantly improves real-world robot performance. 
Our project is at \href{https://diffusion-trajectory-guided-policy.github.io/}{diffusion-trajectory-guided-policy.github.io/}.
\end{abstract}

\section{Introduction}
%
%
%
%

\label{sec:introduction}

\begin{figure*}
\begin{center}
\includegraphics[width=\linewidth]{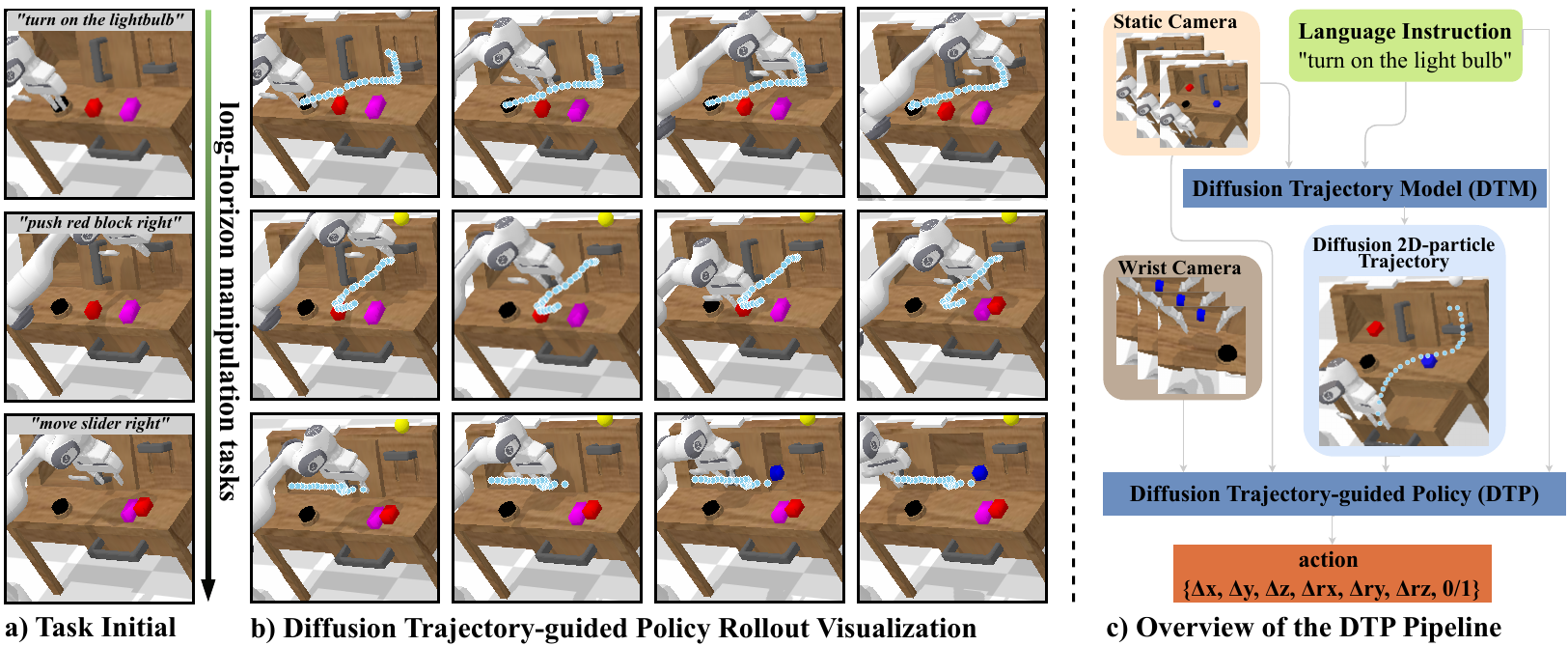} 
\end{center}
\caption{\textbf{System overview.} a) and b) present a task instruction with the initial task observation, allowing our Diffusion Trajectory Model to predict the complete future 2D-particle trajectories; c) illustrates the Diffusion Trajectory-guided pipeline, showcasing how these predicted trajectories guide the manipulation policy.}
\label{fig:overview} 
\vspace{-0.5cm}
\end{figure*}

Imitation Learning (IL) demonstrates significant potential in addressing manipulation tasks within real robotic systems, this is evidenced by its ability to acquire diverse behaviors such as preparing coffee~\cite{zhu2023viola} and flipping mugs~\cite{chi2023diffusionpolicy} through learning from expert demonstrations. 
However, these demonstrations are often limited in coverage~\cite{gao2024prime}, failing to encompass every possible robot pose and environmental variation throughout long-horizon manipulation tasks (Fig.~\ref{fig:overview}a)). This limitation leads to a key challenge in IL—compounding errors over extended trajectories, where small deviations from the expert trajectory accumulate, ultimately causing task failures.

Additionally, robot data is often scarce compared to computer vision tasks because it requires costly and time-consuming human demonstrations. Therefore, improving the generalization capabilities of imitation learning methods using extremely limited and scarce data, given the constraints and high costs of expert demonstrations, becomes a significant challenge. 

Recent research has introduced Vision-Language Action (VLA) models~\cite{brohan2022rt,brohan2023rt,ma2024survey} that map multi-modal inputs to robot actions using Transformer structures~\cite{vaswani2017attention}. Some approaches, like Susie~\cite{black2023zero} and others~\cite{du2023video,du2024learning}, integrate vision and language to generate goal images or future videos, pretrained on large-scale internet datasets. RT-Trajectory~\cite{gu2023rt} uses coarse trajectory sketches instead of language, while RT-H~\cite{belkhale2024rt} breaks down complex instructions into simpler, hierarchical commands. For instance, "Close the pistachio jar" is decomposed into steps like "rotate arm right" and "move the arm forward," facilitating robot action generation.
These methods transform complex instructions into goal images, replace language with trajectory sketches, or simplify instructions into directional commands, mitigating compounding errors in imitation policies, especially in long-horizon tasks. However, they often depend on manually provided trajectories or goal images, limiting flexibility in diverse or unstructured environments.

In this paper, we present a novel diffusion-based paradigm aimed at bridging the feature gap between vision-language inputs and action spaces. By generating task-relevant 2D trajectories from vision-language inputs and mapping them to the action space, our approach improves performance in long-horizon robotic manipulation tasks. Unlike robots that rely on precise instructions, humans use high-level visualizations, like imagined trajectories, to intuitively guide actions, adapting to changes and refining movements in real-time. Similarly, instructing a robot with language should allow envisioning a trajectory to guide future actions based on current observations.
To achieve this, we introduce the Diffusion Trajectory-guided Policy (DTP), consisting of two stages: the Diffusion Trajectory Model (DTM) learning stage and the vision-language action policy learning stage, as depicted in Fig.~\ref{fig:overview}c). The first stage generates a task-relevant trajectory using a diffusion model, which then guides the robot's manipulation policy learning in the second stage, enhancing data efficiency and generalization.
The two-stage design allows the first stage to set conditions for the second. We designed the first stage as an independent helper module, easily integrated into any Transformer-based baseline as an additional input, making our approach adaptable across different models.
We validated our method through extensive experiments on the CALVIN simulation benchmark~\cite{mees2022calvin}, achieving a 25\% higher average success rate than state-of-the-art baselines across various settings. Additionally, our approach is computationally efficient, requiring only consumer-grade GPUs for training.
The main contributions of the paper include:
\begin{enumerate}
    \item We propose the Diffusion Trajectory-guided Policy (DTP), a novel imitation learning framework that utilizes a diffusion trajectory model to guide policy learning for long-horizon robot manipulation tasks.

    \item We leverage robot video data to pretrain a generative vision-language diffusion model, which enhances imitation policy training efficiency by fully utilizing available robot data. Furthermore, our method can be combined with large-scale pretraining methods, serving as a simple and effective plugin to enhance performance.

    \item We conducted extensive experiments in both simulated and real-world environments to evaluate the performance of DTP across diverse settings.
\end{enumerate}

\section{Related Work}
\label{sec:related_work}

\textbf{Language-conditioned Visual Manipulation Policy Control.} Language-conditioned visual manipulation has made significant progress due to advancements in large language models (LLMs) and vision-language models (VLMs).
By using task planners like GPT-4~\cite{achiam2023gpt} or PaLM-E~\cite{driess2023palm}, it is possible to break down complex embodied tasks into simpler, naturally articulated instructions. Recently, several innovative methods have been developed in this domain. RT-1~\cite{brohan2022rt} pioneered the end-to-end generation of actions for robotic tasks. RT-2~\cite{brohan2023rt} explores the capabilities of LLMs for Vision-Language-Action (VLA) tasks by leveraging large-scale internet data. RoboFlamingo~\cite{li2024visionlanguage} follows a similar motivation as RT-2, focusing on the utilization of extensive datasets. RT-X~\cite{o2024open} prioritizes the accumulation of additional robotic demonstration data to refine training and establish scaling laws in robotic tasks. The Diffusion Policy~\cite{chi2023diffusionpolicy} addresses the prediction of robot actions using a denoising model. Lastly, Octo~\cite{octo_2023} serves as a framework for integrating the aforementioned contributions into a unified system, further advancing the field of language-conditioned visual manipulation.

\textbf{Policy Conditioning Representations.}
Leveraging the high-dimensional semantic information in language, video prediction as a pre-training method~\cite{du2024learning,escontrela2024video} yields reasonable results by generating future subgoals for the policy to achieve. Similarly, the goal image generation method~\cite{black2023zero} uses subgoal images instead of full video sequences. However, both approaches often result in hallucinations and unrealistic movements, and they require substantial computational resources, especially during inference.
Methods such as those represented by MimicPlay~\cite{wang2023mimicplay} involve learning a latent planner~\cite{wu2024discrete}. These approaches require an additional training phase, and the latent planner's ability to learn useful features can only be indirectly visualized through a decoder, which is not very intuitive. RT-Trajectory~\cite{gu2023rt} and ATM~\cite{wen2023any} provide innovative methods for generating coarse or particle trajectories. 
Unlike RT-Trajectory, which uses coarse trajectories with significant noise, we use particle trajectories for generation precision and flexibility. In contrast to ATM, which tracking any sampled points, we use a single key point to illustrate the task process regarding the end-effector's position in RGB space. This key point is readily derived using the camera's intrinsic and extrinsic parameters.
Motion Tracks~\cite{ren2025motion} and Im2Flow2Act~\cite{xu2024flow} stem from ATM methods. Our method diverges from Motion Tracks by generating language-related long-horizon 2D points, while Motion Tracks only generates short-horizon 2D points without language input. Im2Flow2Act directly uses 2D flow for policy conditions, losing depth information. In contrast, our framework merges the benefits of each method, using 2D points to boost policy output. To standardize the notion of 2D points or waypoints in the RGB space, we label the key point sequences throughout a task as 2D-particle trajectories.
Our method functions similarly to video prediction, serving as a plugin to enhance policy learning.

\textbf{Diffusion Model for Generation.}
Diffusion models in robotics are primarily utilized in two areas. Firstly, as previously discussed~\cite{black2023zero,du2023video,du2024learning}, they are used for generating future imagery in both video and goal image generation tasks. Secondly, diffusion models are applied to visuomotor policy development, as detailed in recent studies~\cite{chi2023diffusionpolicy, su2025freqpolicy, octo_2023}. These applications highlight the versatility of diffusion models in enhancing robotic functionalities. 
Unlike these methods, our approach uses diffusion models not to directly generate the final policy but to create a 2D-particle trajectory for future end-gripper movement planning in the RGB domain.

\begin{figure*}
\begin{center}
\includegraphics[width=\linewidth]{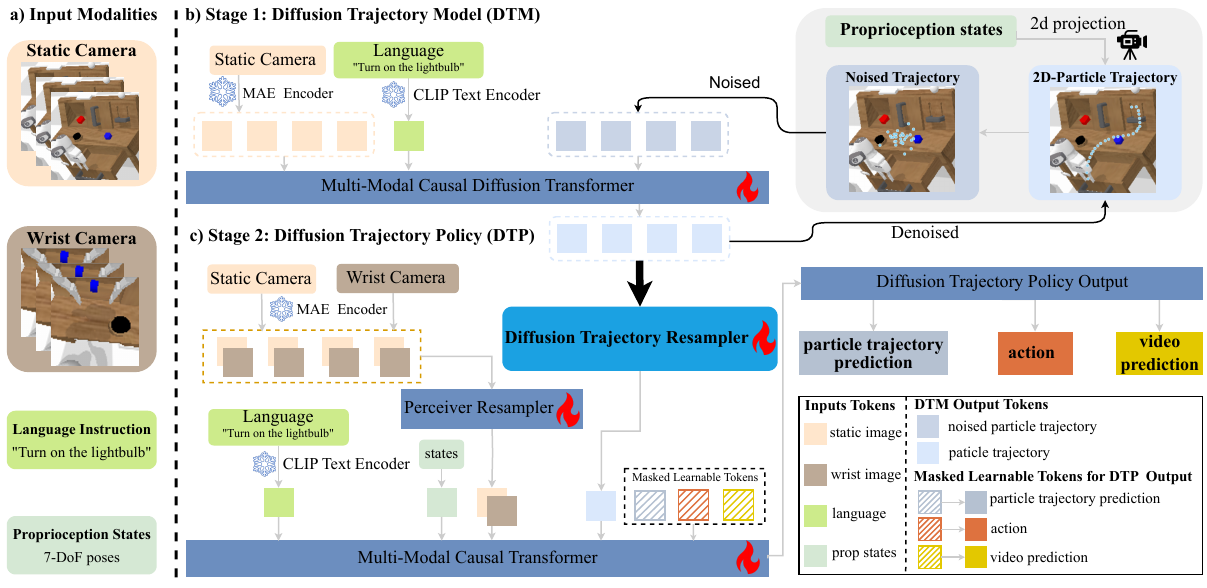} 

\end{center}
\caption{\textbf{System architecture} for learning language-conditioned policies. a) shows the input modalities, including vision, language, and proprioception. b) describes the Diffusion Trajectory Model, detailing how vision and language inputs generate diffusion particle trajectories. c) explains how these trajectories guide the training of robot policies, focusing on the learning of the Diffusion Trajectory Policy. Masked learnable tokens represent the particle trajectory prediction token, action token, and video prediction token, respectively. 
}
\label{fig:framework} 
\vspace{-0.5cm}
\end{figure*}

\section{\uppercase{Method}}
\label{sec:method}


Our goal is to create a policy that enables robots to handle long-horizon manipulation tasks by interpreting vision and language inputs. We simplify the VLA task using two distinct phases (Fig.~\ref{fig:framework}b)c)): a DTM learning phase and a DTP learning phase. First, we generate diffusion-based 2D particle trajectories for the task. Subsequently, in the second stage, these trajectories are used to guide the learning of the manipulation policy.

\subsection{Problem Formulation}
\textbf{Multi-Task Visual Robot Manipulation.} We consider the problem of learning a language-conditioned policy $\pi_\theta$ that takes advantage of language instruction $l$, observation $\vo_t$, robot states $\vs_t$ and diffusion trajectory $\vp_{t:T}$ to generate a robot action $\va_t$:
\begin{equation}
\label{base_policy}	\pi_\theta(l,\vo_t,\vs_t,\vp_{t:T})\rightarrow\\\va_t\
\end{equation}
The robot receives language instructions detailing its objectives, such as "turn on the light bulb". The observation sequence, $\vo_{t-h:t}$, captures the environment's data from the previous $h$ time steps. The state sequence, $\vs_{t-h:t}$, records the robot's configurations, including the pose of the end-effector and the status of the gripper. The diffusion trajectory, $\vp_{t:T}$, predicts the future movement of the end-gripper from time $t$ to the task's completion at time $T$. Our dataset, $\sD$, comprises $n$ expert trajectories across $m$ different tasks, denoted as $\sD_m=\{\tau_i\}_{i=1}^n$. Each expert trajectory $\tau$ includes a language instruction along with a sequence of observation images, robot states, and actions: $\tau = \{\{l,\vo_1,\vs_1,\va_1\}\,...,\{l,\vo_T,\vs_T,\va_T\}\}$.
\subsection{Framework}
\label{Framework}
We introduce the Diffusion Trajectory-guided Policy, as illustrated in Fig.~\ref{fig:framework}. DTP operates within a two-stage framework. In the first stage, our primary focus is on generating the diffusion trajectory $\vp_{t:T}$ which outlines the motion trends essential for completing the task, as observed from a static perspective camera (Fig.~\ref{fig:framework}b)). This 2D-particle trajectory serves as the guidance for subsequent policy learning.
We take a causal Transformer as the backbone network which is designed to handle diverse modalities, processing inputs to predict future images and robotic actions with learnable observation and action query tokens respectively. It integrates CLIP~\cite{radford2021learning} as the language encoder for processing language instructions $l$ and employs a MAE~\cite{he2022masked} as the vision encoder for \(\vo_{t-h:t}\), both of which are with frozen parameters. The vision tokens are then processed with a perceiver resampler~\cite{jaegle2021perceiver} to reduce their number. Additionally, it incorporates the robot's state \(\vs_{t-h:t}\) in world coordinates, as part of its input. All input modalities are shown in Fig.~\ref{fig:framework}a).
Our approach is divided into two main sections. Initially, we detail the process of learning a diffusion trajectory model from the dataset $\sD$ in Section \ref{Diffusion Trajectory Model}. Subsequently, in Section \ref{Diffusion Trajectory-guided Policy}, we illustrate how diffusion trajectories can be used to guide policy learning for long-horizon robot tasks. 
\subsection{Diffusion Trajectory Model}
\label{Diffusion Trajectory Model}
In the first stage (Fig.~\ref{fig:framework}b)), we focus on generating diffusion trajectory that maps out the motion trends required for task completion, as viewed from a static perspective camera. To achieve this, we employ a model $\mM_d$ to transform language instructions $l$ and initial visual observations $\vo_t$ into a sequence of diffusion 2D-particle trajectories \(\vp_{t:T}\). These points indicate the anticipated movements for the remainder of the task:
\begin{equation}
	\mM_d(l,{\vo_t})\rightarrow\\\vp_{t:T}\
\label{eq:diff}
\end{equation}
\subsubsection{Data Preparation}
According to Eq. \ref{eq:diff}, our input consists of observations $\vo_t$ and language instruction $l$. For outputs, our aim is to determine the future 2D-particle trajectory $\vp_{t:T}$ of the end effector gripper for finishing the task. Recent advancements in video tracking work make it easy to monitor the end effector gripper~\cite{yang2023track}. For enhanced convenience and precision, we achieve this by mapping the world coordinates $(x_w,y_w,z_w)$ to pixel-level positions $(x_c,y_c)$ according to camera’s intrinsic and extrinsic parameters in the static camera frame, as shown in (Fig.~\ref{fig:framework}b)) right part. In the first stage, our data format is structured as \(\sD_{\text{trajectory}} = \{l, \vo_t, \vp_{t:T}\}\), facilitating straightforward acquisition of the sequence \(\vp_{t:T}\), thereby simplifying the process of training our model to accurately predict end effector positions.
\subsubsection{Training Objective}
Denoising Diffusion Probabilistic Models (DDPMs)~\cite{ho2020denoising} constitute a class of generative models that predict and subsequently remove noise during the generation process. In our approach, we utilize a causal diffusion decoding structure~\cite{chi2023diffusionpolicy} to generate diffusion 2D-particle trajectories $\vp_{t:T}$. Specifically, we initiate the generation process by sampling a Gaussian noise vector $x^K \sim \mathcal{N}(0, I)$ and proceed through $K$ denoising steps using a learned denoising network \( \epsilon_\theta(x^k, k) \) where $x^k$ represents the diffusion trajectory noised over $K$ steps. This network iteratively predicts and removes noise $K$ times, ultimately resulting in the output $x^0$, which denotes the complete removal of noise. The process is described in the equation below, where $\alpha$, $\gamma$, and $\sigma$ are parameters that define the denoising schedule:
\begin{equation}
x^{k-1} = \alpha (x^k - \gamma \epsilon_\theta (x^k, k)) + \mathcal{N}(0, \sigma^2 I)
\label{eq:base_diff}
\end{equation}
Eq. \ref{eq:base_diff}, illustrates the functioning of the basic diffusion model. For our application, we adapt this model to generate diffusion trajectories $\vp_{t:T}$ based on the observation $\vo_t$ and language instruction $l$:
\begin{equation}
\vp_{t:T}^{k-1} = \alpha (\vp_{t:T}^k - \gamma \epsilon_\theta (\vo_t,l,\vp_{t:T}^k, k)) + \mathcal{N}(0, \sigma^2 I)
\label{eq:2d_diff}
\end{equation}
During the training process, the loss is calculated as Mean Square Error (MSE), where $\epsilon_k$ represents Gaussian noise sampled randomly for step $k$:
\begin{equation}
\mathcal{L}_{DTM} = \text{MSE}(\epsilon_k, \epsilon_\theta(\vo_t, l, \vp_{t:T} + \epsilon_k, k))
\end{equation}

This transformation integrates our specific inputs into the diffusion process, enabling the tailored generation of diffusion trajectory in alignment with both the observed data and the provided language instruction.
This training loss ensures that diffusion 2D-particle trajectories are accurately generated by systematically reducing noise, thereby enhancing the precision of the final trajectory predictions.

\subsection{Diffusion Trajectory-guided Policy}
\label{Diffusion Trajectory-guided Policy}
In the second stage, we focus on illustrating how the diffusion trajectory guides the robot manipulation policy (Fig.~\ref{fig:framework}c)). As previously outlined in our problem formulation, we define our task as a language-conditioned visual robot manipulation task. We base our Diffusion Trajectory-guided Policy on the GR-1~\cite{wu2023unleashing} model and incorporate our diffusion trajectory $\vp_{t:T}$ as an additional input, as specified in Eq. \ref{base_policy}.	

\textbf{Policy Input.} This consists of language and image inputs, as detailed in the Sec.~\ref{Framework} and shown in the left side of 
Fig.~\ref{fig:framework}c). To clearly demonstrate our method's performance, we maintain the same configuration as GR-1.
Importantly, for the diffusion trajectory, we do not rely on the inference results from the first training stage. Instead, we use the labeled data from this stage as the diffusion trajectory. This approach enhances precision in training and conserves computational resources by using the labels directly. The simplest training approach is to inject the diffusion particle trajectory directly into the causal baseline. However, our fixed set of 2D particle trajectories $\vp_{t:T}$ can lead to computational intensity during training due to the high number of tokens. Inspired by the perceiver resampler~\cite{jaegle2021perceiver}, we designed a diffusion trajectory resampler module to reduce the number of trajectory tokens, as shown in Fig.~\ref{fig:framework}b) and c). 

\textbf{Diffusion Trajectory-guided Policy Training.} During the policy learning phase (Fig.~\ref{fig:framework}c)), we generate future particle trajectories 
to supervise the diffusion trajectory resampler module with \(\mathcal{L}_{\text{trajectory}}\). Our policy framework employs a causal Transformer architecture, where future particle trajectory tokens are generated prior to action tokens with \(\mathcal{L}_{\text{action}}\).  This ensures that the particle trajectory tokens effectively guide the formation of action tokens, optimizing the action prediction process in a contextually relevant manner. Additionally, we retain the output of video prediction with \(\mathcal{L}_{\text{video}}\), maintaining the same setting as GR-1. This consistency in output makes it easier to conduct ablation studies, as we can directly compare our approach to the original GR-1 model. The optimal DTP objective can be expressed as the following equation:

\begin{equation}
\mathcal{L}_{DTP} = \mathcal{L}_{trajectory} + \mathcal{L}_{action} + \mathcal{L}_{video}
\end{equation}

Furthermore, to demonstrate the effectiveness and superiority of our method in the ablation study, we split the GR-1 baseline into two versions: one that is fully pretrained on the video dataset and another that only uses the GR-1 structure without any pretraining. We will discuss these two baseline configurations in Sec.~\ref{sec:experiment}.

\section{\uppercase{Experiment}}
\label{sec:experiment}


In this section, we evaluate the performance of Diffusion Trajectory Policy in both simulation and real-world robot experiments. 
\subsection{CALVIN Benchmark and Baselines}

\begin{figure}[t]
\begin{center}
\includegraphics[width=\linewidth]{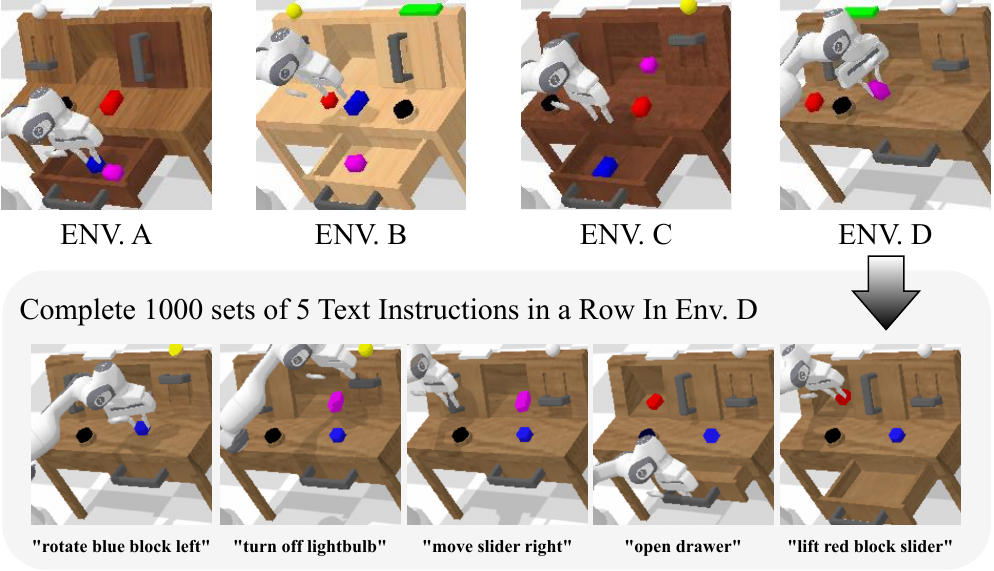} 
\end{center}
\caption{The upper four environments correspond to the CALVIN ABCD settings. The bottom section shows a sequence of five long-horizon tasks, each guided by a specific instruction.}
\label{fig:env} 
\vspace{-0.5cm}
\end{figure}

CALVIN~\cite{mees2022calvin} is a comprehensive simulated benchmark designed for evaluating language-conditioned policies in long-horizon robot manipulation tasks. It comprises four distinct yet similar environments (A,B,C, and D) which vary  in desk shades and item layouts, as shown in Fig.~\ref{fig:env}. This benchmark includes 34 manipulation tasks with unconstrained language instructions. Each environment features a Franka Emika Panda robot equipped with a parallel-jaw gripper, and a desk that includes a sliding door, a drawable drawer, color-varied blocks, an LED, and a light bulb, all of which can be interacted with or manipulated.

\textbf{Experiment Setup.} 
We train DTP to predict relative actions in $xyz$ positions and Euler angles for arm movements, along with binary actions for the gripper. Our simulation setup aligns with the base model~\cite{wu2023unleashing}, where both the number of generated and executed actions are set to 1. The training dataset includes over 20,000 expert trajectories from four scenes, each paired with language instruction labels. Our DTP method is evaluated using the long-horizon benchmark, which features 1,000 unique sequences of instruction chains articulated in natural language. Each sequence requires the robot to sequentially complete five tasks.

\textbf{Baselines.} We compare our proposed policy against the following state-of-the-art language-conditioned multi-task policies on CALVIN: 
\textbf{MT-ACT}~\cite{bharadhwaj2024roboagent}: A multitask Transformer-based policy predicting action chunks.
\textbf{HULC}~\cite{mees2022matters}: A hierarchical approach predicting latent sub-goal features from language and observations.
\textbf{RT-1}~\cite{brohan2022rt}: Utilizes convolutional layers and Transformers for end-to-end action generation from language and observations.
\textbf{RoboFlamingo}~\cite{li2023vision}: A fine-tuned Vision-Language Foundation model with 3 billion parameters.
\textbf{GR-1}: Pretrained on the Ego4D dataset, featuring large-scale human-object interactions.
\textbf{3D Diffuser Actor}~\cite{ke20243d}: Integrates 3D scene representations with diffusion objectives to learn policies from demonstrations.
\begin{table}[t]
\centering
\caption{Summary of Experiments}
\label{tab:experiment_summary} 
\resizebox{0.5\textwidth}{!}{ 
\begin{tabular}{c|c|ccccc|c}
\toprule

\textbf{Method} & \textbf{Experiment} & \multicolumn{5}{c}{\textbf{Tasks completed in a row}} & \textbf{Avg. Len.} \\
\cmidrule(lr){3-7}
& & \textbf{1} & \textbf{2} & \textbf{3} & \textbf{4} & \textbf{5} & \\
\midrule
HULC     & D$\rightarrow$D & 0.827 & 0.649 & 0.504 & 0.385 & 0.283 & 2.64 \\
GR-1     & D$\rightarrow$D & 0.822 & 0.653 & 0.491 & 0.386 & 0.294 & 2.65 \\
MT-ACT   & D$\rightarrow$D & 0.884 & 0.722 & 0.572 & 0.449 & 0.353 & 3.03 \\
HULC++    & D$\rightarrow$D & 0.930 & 0.790 & 0.640 & 0.520 & 0.400 & 3.30 \\
\rowcolor{gray!30} DTP (ours) & D$\rightarrow$D & 0.924 & 0.819 & 0.702 & 0.603 & 0.509 & 3.55 \\

\midrule
HULC             & ABC$\rightarrow$D & 0.418 & 0.165 & 0.057 & 0.019 & 0.011 & 0.67 \\
RT-1             & ABC$\rightarrow$D & 0.533 & 0.222 & 0.094 & 0.038 & 0.013 & 0.90 \\
RoboFlamingo     & ABC$\rightarrow$D & 0.824 & 0.619 & 0.466 & 0.380 & 0.260 & 2.69 \\
GR-1             & ABC$\rightarrow$D & 0.854 & 0.712 & 0.596 & 0.497 & 0.401 & 3.06 \\
3D Diffuser Actor & ABC$\rightarrow$D & 0.922 & 0.787 & 0.639 & 0.512 & 0.412 & 3.27 \\
\rowcolor{gray!30} DTP (ours) & ABC$\rightarrow$D & 0.890 & 0.773 & 0.679 & 0.592 & 0.497 & 3.43 \\
\midrule
RT-1             & 10\% ABCD$\rightarrow$D &0.249&0.069&0.015&0.006&0.000&0.34 \\ 
HULC             & 10\% ABCD$\rightarrow$D &0.668&0.295&0.103&0.032&0.013&1.11 \\  
GR-1             & 10\% ABCD$\rightarrow$D &0.778&0.533&0.332&0.218&0.139&2.00 \\ 
\rowcolor{gray!30} DTP (ours)     & 10\% ABCD$\rightarrow$D & 0.813 & 0.623 & 0.477 & 0.364 & 0.275 & 2.55 \\ 

\bottomrule
\end{tabular}
}
\caption*{\footnotesize This table details the performance of all baseline methods in sequentially completing 1, 2, 3, 4, and 5 tasks in a row. The average length, shown in the last column and calculated by averaging the number of completed tasks in a series of 5 across all evaluated sequences, illustrates the models' long-horizon capabilities. 10\% ABCD$\rightarrow$D indicates that only 10\% of the training data is used.}
\vspace{-0.5cm}
\end{table}

\subsection{Comparisons with State-of-the-Art Methods}
\textbf{Results in Seen Environments.} In the D$\rightarrow$D setting, using about 5,000 expert demonstrations, training on 8*3090 GPUs took 1.5 days. As shown in Table~\ref{tab:experiment_summary}, DTP outperforms all baselines in long-horizon tasks, increasing Task 5's success rate from 0.400 to 0.509 and the average sequence length from 3.30 to \textbf{3.55}. Compared to GR-1, DTP boosts performance across all metrics, with a 33.9\% increase in average sequence length, demonstrating superior performance as task length increases.

\textbf{Results in Unseen Environments.} In the challenging ABC$\rightarrow$D setting, models are trained on environments A, B, and C, and tested in the unseen environment D. Training took about 5 days on 8*3090 GPUs. As shown in Table~\ref{tab:experiment_summary}, DTP demonstrates strong generalization, increasing the average task completion length from 3.06 to \textbf{3.43} and achieving a Task 5 success rate of 0.497, the highest recorded. Notably, without using depth modality, DTP outperformed the 3D Diffuser Actor, highlighting its effectiveness in guiding policy learning for long-horizon tasks in unseen settings.

\textbf{Data Efficiency.} In the ABCD$\rightarrow$D setting, we evaluated data efficiency by training with only 10\% of the dataset, using around 2,000 expert demonstrations. Training took about 1 day on 8*3090 GPUs. As shown in Table~\ref{tab:experiment_summary}, while all methods perform worse with less data, the best baseline, GR-1, achieves a success rate of 0.778 with an average length of 2.00. DTP excels in long-horizon tasks, with a success rate increase and an average length of \textbf{2.55}, outperforming others. 
These results demonstrate the data efficiency of DTP. By leveraging diffusion-based trajectories, the policy effectively captures positional preferences that are critical for long-horizon tasks. Furthermore, these 2D-particle trajectories provide guidance to the robot arm, enabling it to acquire skills even with a limited number of demonstrations.

\subsection{Ablation Studies}
\label{Ablation Studies}
In this section, we conduct ablation studies to assess how diffusion trajectories enhance policy learning in visual robot manipulation tasks. This key contribution significantly improves imitation policy training efficiency by fully utilizing robot data. Integrated with large-scale pretraining baselines, it offers a straightforward performance boost. We compare our method against two baselines: one using the GR-1 framework (Sec.~\ref{Framework}) without video pretraining, and another with large-scale video pretraining using the Ego4D~\cite{grauman2022ego4d} dataset, both based on GR-1. These baselines verify our method's efficacy and compatibility.

\begin{table}[t]
\centering
\caption{Ablation Studies}
\label{tab:Ablation Studies} 
\resizebox{0.485\textwidth}{!}{ 
\begin{tabular}{cc|c|ccccc|c}
\toprule
\textbf{Pre-Training}  & \textbf{DTP (Ours)} & \textbf{Data} & \textbf{1} & \textbf{2} & \textbf{3} & \textbf{4} & \textbf{5} & \textbf{Avg. Len.} \\
\midrule

$\times$  & $\times$  & ABC$\rightarrow$D & 0.815 & 0.651 & 0.498 & 0.392 & 0.297 & 2.65 \\
\rowcolor{gray!30}$\times$  & \checkmark  & ABC$\rightarrow$D & 0.869 & 0.751 & 0.636 & 0.549 & 0.465 & 3.27 \\
$\times$& $\times$ &10\% ABCD$\rightarrow$D &0.698&0.415&0.223&0.133&0.052&1.52\\
\rowcolor{gray!30}$\times$& \checkmark &10\% ABCD$\rightarrow$D &0.742&0.511&0.372&0.269&0.188&2.08\\

\midrule

\checkmark  & $\times$ & ABC$\rightarrow$D & 0.854 & 0.712 & 0.596 & 0.497 & 0.401 & 3.06 \\
\rowcolor{gray!30}\checkmark  & \checkmark & ABC$\rightarrow$D & 0.890 & 0.773 & 0.679 & 0.592 & 0.497 & 3.43 \\
\checkmark& $\times$ &10\% ABCD$\rightarrow$D &0.778&0.533&0.332&0.218&0.139&2.00\\  
\rowcolor{gray!30}\checkmark & \checkmark &10\% ABCD$\rightarrow$D&0.813 & 0.623 & 0.477 & 0.364 & 0.275 & 2.55 \\ 
\rowcolor{gray!30}\checkmark & 100\%\checkmark &10\% ABCD$\rightarrow$D& 0.822 & 0.643 & 0.526 & 0.416 & 0.302 & 2.71 \\ 

\bottomrule
\end{tabular}
}

\caption*{\footnotesize Pre-Training indicates whether we use only the baseline model structure or the baseline pre-trained on the Ego4D dataset. DTP (ours) indicates whether the generated 2D particle trajectory is input into the policy. In our ablation studies, we established these two baselines to evaluate the effectiveness and compatibility of our DTM method with other approaches. 10\% ABCD$\rightarrow$D indicates that only 10\% of the training data is used. 100\%\checkmark indicates DTM trained on full ABCD$\rightarrow$D.}
\vspace{-0.6cm}
\end{table}

\textbf{Diffusion Trajectory Policy from Scratch.} 
We evaluate our method in the ABC$\rightarrow$D and 10\% ABCD$\rightarrow$D settings (see Table~\ref{tab:Ablation Studies}). The diffusion trajectory method significantly enhances performance without pretraining, excelling in sequential tasks and increasing the average task completion length by 23.4\%. The success rate for Task 5, indicative of long-horizon success, rises by 56.6\%.

\textbf{Diffusion Trajectory Policy with Video Pretrain.} 
As shown in the bottom part of Table~\ref{tab:Ablation Studies}, our diffusion trajectory variants enhance baseline model performance to state-of-the-art levels. Evaluated in ABC$\rightarrow$D and 10\% ABCD$\rightarrow$D settings, our method consistently outperforms traditional scratch training, significantly boosting baseline performance. Success rates increase notably, from 4.2\% in the first task to 23.9\% in the fifth, validating DTP's effectiveness in long-horizon manipulation tasks.

\textbf{Diffusion Trajectory Model Scaling Law.} 
The last row highlights the initial training stage of our Diffusion Trajectory Model. Increasing training data improves point accuracy, enhancing DTP. Even with limited demonstration data, scaling diffusion trajectory training boosts success rates and task completion length. This suggests a potential direction: while robot demonstration data is costly, DTM data is easier to annotate, requiring only a coarse trajectory sketch on an RGB image with language instructions.

\begin{figure*}
\begin{center}
\includegraphics[width=\linewidth]{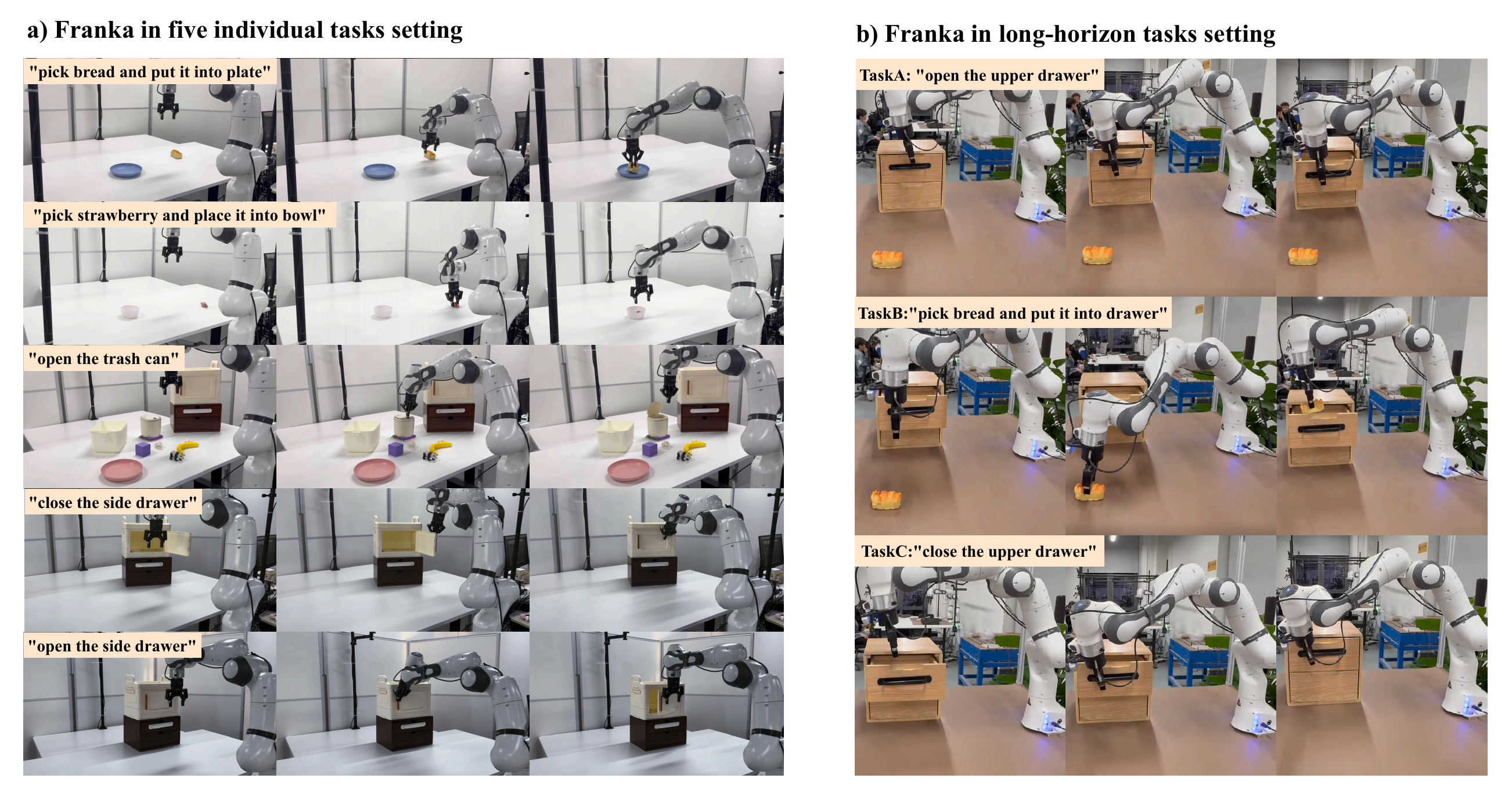} 
\end{center}
\caption{Real-robot experiments: a) Franka performing five distinct manipulation tasks. b) Franka performing one long-horizon task composed of subtasks (A–B–C).}
\label{fig:real_robot_summary} 
\vspace{-0.5cm}
\end{figure*}

\begin{figure}
\begin{center}
\includegraphics[width=\linewidth]{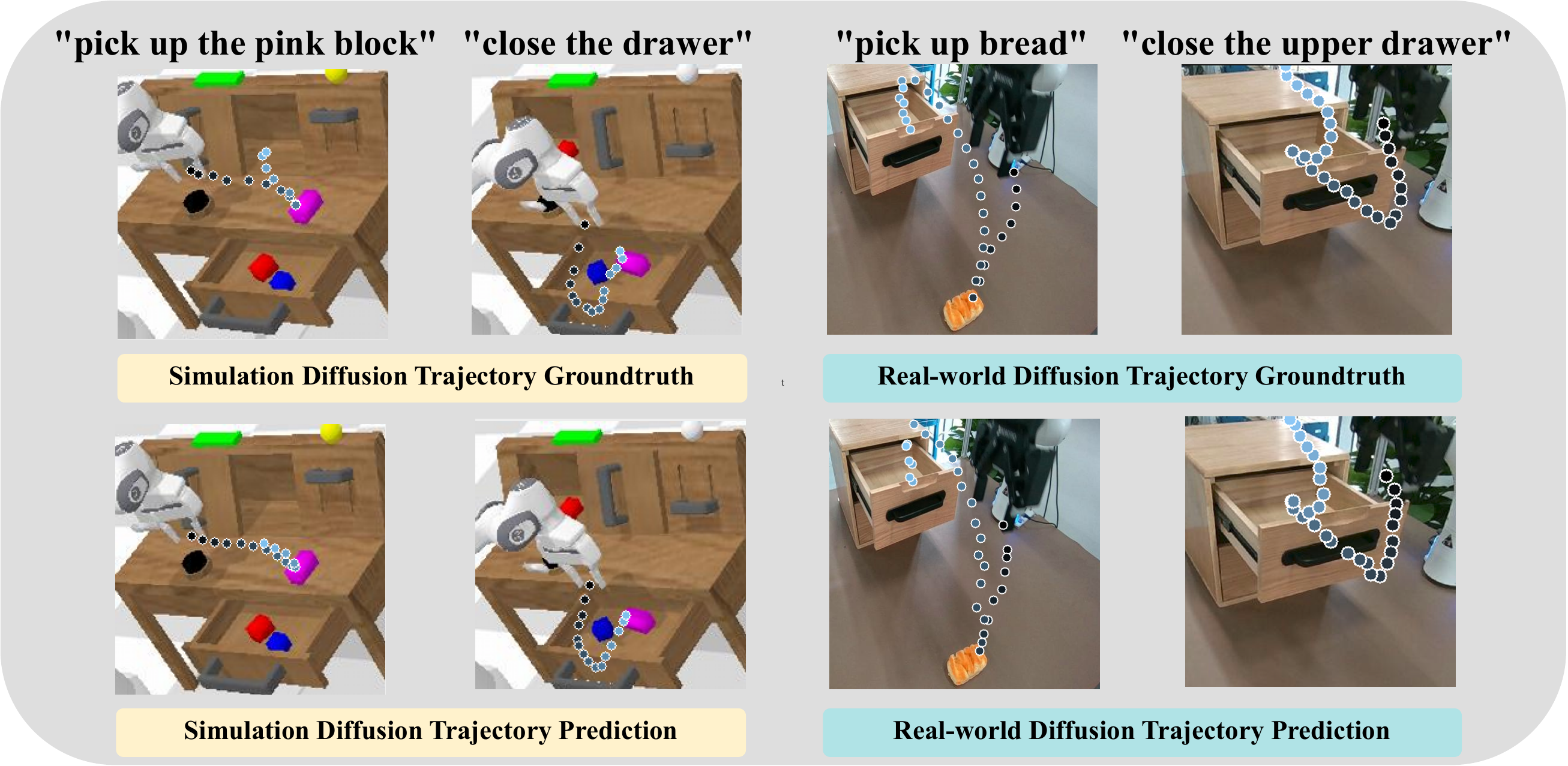} 
\end{center}
\caption{\textbf{Diffusion Trajectory Visualization.} The left half part illustrates diffusion trajectory generation in the CALVIN environment, while the right half part show trajectory generation in a real-world robotic scenario.}
\label{fig:visualization} 
\vspace{-1cm}
\end{figure}

\subsection{Real Robot Experiment}


\textbf{Experiment Setup.} Our robotic system features a Franka Emika Panda robot with three Intel RealSense D435i cameras and a Robotiq gripper.
We collected 1,512 demonstrations using a teleoperation system~\cite{wu2024gello}, with 290, 258, 100, 184, and 254 demonstrations for the tasks \textit{PickBread}, \textit{PickStrawberry}, \textit{OpenTrash}, \textit{CloseSideDrawer}, and \textit{OpenSideDrawer}, respectively, covering object transportation and manipulation (see Fig.~\ref{fig:real_robot_summary}a). Additionally, we collected 426 demonstrations for a long-horizon task comprising three subtasks(A-B-C) (see Fig.~\ref{fig:real_robot_summary}b).
In real robot setup, the number of generated actions is set to 25. This adjustment addresses sim-to-real discrepancies and latency, as setting both values to 1 causes the robot to remain almost completely still, with only minor vibrations. The number of executed actions remains at 1. We also incorporate an action ensemble mechanism, similar to ALOHA~\cite{zhao2023learning}. The training process spanned 20 epochs and took approximately one day on four RTX 3090 GPUs.

\textbf{Results.} The performance of DTP and baseline methods in five individual tasks is summarized in Table~\ref{tab:real_robot_results} upper part. Each task was evaluated over 10 trials, with success rates calculated for comparison. Overall, DTP achieved the highest aggregate success rate across tasks. However, in the \textit{PickStrawberry} task, DTP underperformed compared to ACT. We attribute this to the small size of the target object, as DTP uses an image input resolution of 224x224, while ACT operates at a higher resolution of 480x640, which likely impacts performance. In long-horizon tasks, the robot arm’s initial pose is determined by the completion of the previous task, resulting in random starting configurations. To evaluate DTP’s robustness in such scenarios, we tested it on the \textit{OpenSideDrawer} task with randomized initial arm poses. DTP achieved a success rate twice as high as the second-best method. Additionally, in the \textit{OpenTrash} task, which requires precise alignment to a specific area to open the trash bin, DTP demonstrated superior guidance capabilities. While other baseline methods positioned the arm near the target, they often failed to locate the precise opening mechanism, leading to task failure.

In Table~\ref{tab:real_robot_results} (bottom), we evaluate variations of the training scenario. While training followed the A–B–C sequence, testing extended to longer sequences of up to five subtasks, such as A-B-C-A-C. This design probes long-horizon challenges: for example, repeating Task A after A-B-C occurs in a new context because the bread is already in the drawer, creating situations unseen during training. We created five such long-sequence variants for evaluation.
Each value in Table~\ref{tab:real_robot_results} indicates the number of successful subtasks (out of five) for our method and the baseline, with “Ave. Len.” reporting the average sequence length achieved. In A-B-C-A-C, the baseline fails at the fourth subtask due to missing bread, while our method continues. In A-C-A-B-C, the direct A→C transition introduces both an unseen scenario and initial pose, limiting the baseline to the first task, whereas our method progresses further. Overall, our approach reaches an average length of 4.6, compared to 2.0 for the baseline. This evaluation highlights the impact of \textbf{\textit{accumulated compounding errors}} in long-horizon tasks: as sequences lengthen and subtask orders vary, accumulated deviations—such as altered object states or unseen robot poses—lead the baseline to fail, while our method remains more robust.

\begin{table}
	\caption{Summary of Real Robot Experiments}
    \resizebox{0.485\textwidth}{!}{ 
	\begin{tabular}{l|ccccc|c}
		\toprule
		\diagbox [width=5em,trim=l] {Method}{Tasks} & \makecell{Pick\\Bread} & \makecell{Pick\\Strawberry} & \makecell{Open\\Trash}   & \makecell{CloseSide\\Drawer}  & \makecell{OpenSide\\Drawer*} & Ave. Suc.  \\
		\midrule
		ACT~\cite{zhao2023learning}       & 0.7 & 0.9 & 0.3 & 0.3 & 0.4 & 0.52 \\
		BAKU~\cite{haldar2024baku}      & 0.0 & 0.5 & 0.2 & 0.2 & 0.3 & 0.24 \\
		GR1~\cite{wu2023unleashing}       & 0.7 & 0.7 & 0.2 & 0.4 & 0.4  &0.48\\
        \rowcolor{gray!30} DTP (ours) & 0.8 & 0.8 & 0.9 & 0.9 & 0.8 & {\bf 0.84} \\
        \midrule
        \diagbox [width=5em,trim=l] {Method}{Tasks} & ABCAC  & ACABC  & CABCA   & CACAB  & BCACA  &  Ave. Len.  \\
        \midrule
        GR1       & 3/5 & 1/5 & 4/5 & 2/5 & 0/5 & 2.0  \\
        \rowcolor{gray!30} DTP (ours) & 5/5 & 5/5 & 5/5 & 3/5 & 5/5  & {\bf 4.6}  \\
	\bottomrule
	\end{tabular}\vspace{0cm}
    }
\caption*{\footnotesize The upper table shows real-robot success rates for five tasks, where OpenSideDrawer* starts from a random robot pose. The lower table extends the long-sequence task A-B-C into longer forms (e.g., A-B-C-A-C) and rearranges them into new variants. Reported values indicate how many of the five subtasks were completed in each variant. "Ave. len" denotes the average sequence length for our method and the baseline GR1.}

\label{tab:real_robot_results}
\vspace{-1.0cm}
\end{table}

\textbf{Visualization of Diffusion Trajectory Model.} As shown in Fig.~\ref{fig:visualization}, we present the overall visualization of the diffusion trajectory generation phase, tested in both the CALVIN environment and real-world scenarios. The visualizations demonstrate that the trajectories generated by our diffusion trajectory prediction closely align with the ground truth. Even when minor deviations occur, the generated trajectories remain consistent with the robotic arm paths dictated by the language instructions.

\section{\uppercase{Conclusion}}
\label{sec:conclusion}
The limited availability of robot data and compounding errors in IL make it difficult to generalize long-horizon tasks to unseen poses and environments. We introduce a diffusion trajectory-guided framework that leverages RGB-domain diffusion trajectories to improve policy learning in robot manipulation. Our method augments training data through data augmentation or manually crafted labels, producing more accurate trajectories. It has two stages: (i) training a diffusion trajectory model to generate task-relevant trajectories, and (ii) using them to guide the robot’s manipulation policy. On the CALVIN benchmark, our method outperforms state-of-the-art baselines by an average success rate of 25\%, and also shows strong improvements using only robot data as well as in real-world experiments.

For future work, we plan to extend our framework to other state-of-the-art policies, as we believe diffusion trajectories can further enhance their effectiveness. Another direction is to obtain trajectory labels using camera intrinsic and extrinsic parameters, which are often missing in open-source datasets~\cite{padalkar2023open}. Track-Anything~\cite{yang2023track} offers strong object-tracking ability for generating labels, while EgoMimic~\cite{kareer2024egomimic} captures detailed 3D hand tracks with Aria glasses that can be projected into 2D particle trajectories. Such hardware also enables large-scale video pretraining for diffusion trajectory tasks.



\bibliographystyle{template/IEEEtran}
\bibliography{DTP}
\end{document}

%% file: template/math_commands.tex
\usepackage{amsmath,amsfonts,bm}

\def\eqref#1{equation~\ref{#1}}

\def\1{\bm{1}}

\def\va{{\bm{a}}}

\def\vo{{\bm{o}}}
\def\vp{{\bm{p}}}

\def\vs{{\bm{s}}}

\def\mM{{\bm{M}}}

\DeclareMathAlphabet{\mathsfit}{\encodingdefault}{\sfdefault}{m}{sl}
\SetMathAlphabet{\mathsfit}{bold}{\encodingdefault}{\sfdefault}{bx}{n}

\def\sD{{\mathbb{D}}}

